\documentclass[10pt,twocolumn,letterpaper]{article}

\usepackage[pagenumbers]{iccv} 

\usepackage{amssymb}  
\usepackage{pifont}   
\usepackage{amssymb}  
\usepackage{pifont}   
\usepackage{colortbl} 
\usepackage{booktabs} 
\usepackage{amsmath}
\usepackage{mathtools}
\usepackage{xcolor}
\usepackage{tikz}
\usepackage{subcaption}
\usetikzlibrary{positioning}
\usepackage{algorithm}
\usepackage{algpseudocode}
\usepackage{setspace}

\definecolor{iccvblue}{rgb}{0.21,0.49,0.74}
\usepackage[pagebackref,breaklinks,colorlinks,allcolors=iccvblue]{hyperref}
\usepackage{graphicx}
\usepackage{amsmath}
\usepackage{amssymb}
\usepackage{colortbl}
\usepackage{graphicx}
\usepackage{xcolor}
\usepackage{multirow}
\usepackage{booktabs} 
\usepackage{amssymb}
\usepackage{pifont}
\usepackage{algorithm}
\usepackage{algpseudocode}

\def\paperID{6882} 
\def\confName{ICCV}
\def\confYear{2025}

\title{Class-Independent Increment: An Efficient Approach for Multi-label
Class-Incremental Learning}

\author{{Chenhao Ding\textsuperscript{\rm 1}} \quad
Songlin Dong\textsuperscript{\rm 3,}\thanks{Corresponding author} \quad 
Zhengdong Zhou\textsuperscript{\rm 1} \quad 
Jizhou Han\textsuperscript{\rm 2} \quad 
Qiang Wang \textsuperscript{\rm 2} \quad \\
Yuhang He\textsuperscript{\rm 2}   \quad
Yihong Gong\textsuperscript{\rm 1,} \textsuperscript{\rm 2}\quad 
\vspace{0.2em} \\
\textsuperscript{\rm 1}School of Software Engineering, Xi’an Jiaotong University \\
\textsuperscript{\rm 2}College of Artificial Intelligence, Xi’an Jiaotong University \\
\textsuperscript{\rm 3}Faculty of Computility Microelectronics, Shenzhen University of Advanced Technology \\
}

\begin{document}
\maketitle
\begin{abstract}
Current research on class-incremental learning primarily focuses on single-label classification tasks. However, real-world applications often involve multi-label scenarios, such as image retrieval and medical imaging. Therefore, this paper focuses on the challenging yet practical multi-label class-incremental learning (MLCIL) problem. In addition to the challenge of catastrophic forgetting, MLCIL encounters issues related to feature confusion, encompassing inter-session and intra-feature confusion. To address these problems, we propose a novel MLCIL approach called class-independent increment (CLIN). Specifically, 
in contrast to existing methods that extract image-level features, we propose a class-independent incremental network (CINet) to extract multiple class-level embeddings for multi-label samples. It learns and preserves the knowledge of different classes by constructing class-specific tokens. On this basis, we develop a novel loss function, optimizing the learning of class-specific embeddings. The loss aims to distinguish between new and old classes, further alleviating the problem of feature confusion. Extensive experiments on MS-COCO and PASCAL VOC datasets demonstrate the effectiveness of our method for improving recognition performance and mitigating forgetting on various MLCIL tasks.

\end{abstract}
\section{Introduction}
\label{sec:intro}
In the past decade, deep learning has catalyzed significant breakthroughs in computer vision and multimedia tasks, including image classification~\cite{song2018deep}, semantic segmentation~\cite{jing2019coarse}, and object detection~\cite{deng2020global}, among others.
However, the deep learning-based methods fail to recognize new samples of unseen classes. Directly learning these new samples will lead to \textit{catastrophic forgetting}~(CF), where the model's performance on historical data significantly deteriorates. In this case, class-incremental learning~(CIL)~\cite{ICARL,l2p} is extensively studied to update the model only with new class data in each task, while preserving the knowledge of old classes.
However, in practical applications, an image typically contains multiple objects. For example, photos on social media often contain multiple labels or objects, making traditional CIL methods that assume each image contains only one object unsuitable. Therefore, some work explores studying CIL in the multi-label scenario, named \textit{multi-label class-incremental learning} (MLCIL)~\cite{KRT,ML-SK,PRS}. It aims to gradually learn new classes from label-incomplete multi-label data streams in different learning stages while maintaining performance on old labels.

Compared to the traditional~(\textit{i.e.} single-label) CIL problem, it is more challenging as a single image may contain multiple objects of both the old and new classes, with only new class objects~(one or multiple categories) annotated in each learning session. To this end, prior methods to address forgetting for MLCIL tasks include exemplar replay (ER), knowledge distillation (KD), and network expansion~(NE). ER-based methods~\cite{PRS} solve the CF problem by storing a small number of old class exemplars. They replay the exemplars with the current session samples to mitigate the old class forgetting when learning new classes. KD-based methods~\cite{KRT,rebll2025} design regularization terms on the image-level feature to preserve previous knowledge when training the model on new data. The NE-based approaches~\cite{KRT,csc,dualprompt} allocate a distinct set of parameters (e.g., tokens, prompts) for each session to learn task-level features. During the training process, parameters associated with old classes irrelevant to the current session are frozen to prevent the forgetting of old class knowledge.

\begin{figure}[t]
    \centering 
    \setlength{\abovecaptionskip}{-0.15cm} 
\setlength{\belowcaptionskip}{-0.25cm} 
    \includegraphics[width=\linewidth]{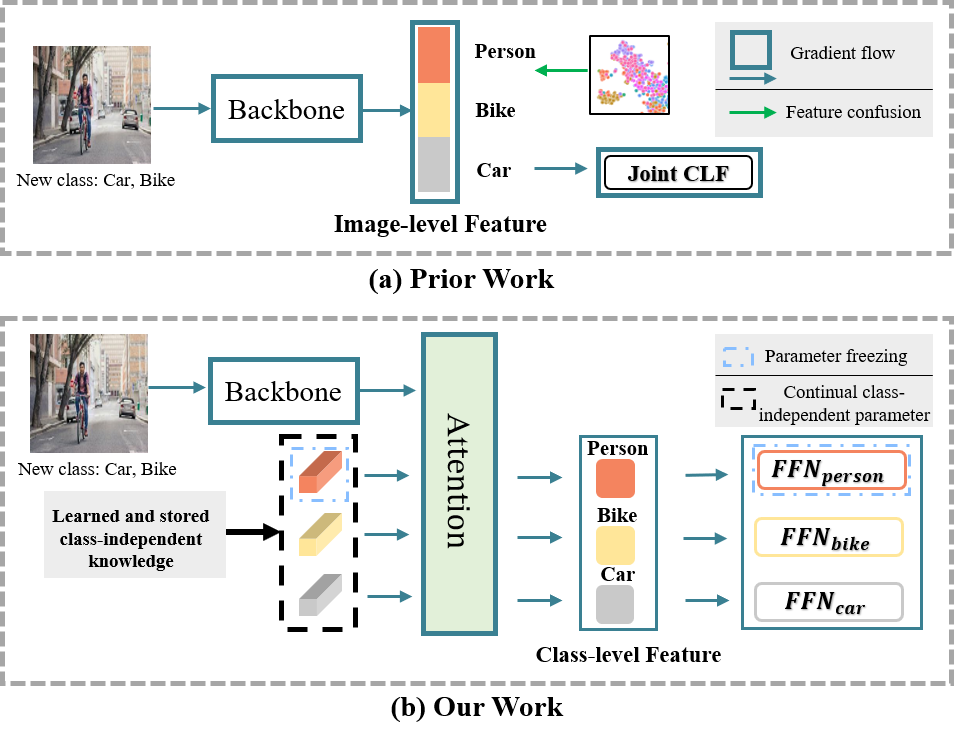}
    \caption{Comparison between \textbf{prior work} and \textbf{our work} (a) The prior paradigm for MLCIL. Prior methods employ image-level (task-level) features and train a joint classifier for each session. (b) Our proposed paradigm. We employ a CLIN framework to generate and process class-level features to solve the feature confusion problem. Moreover, we design two loss functions to prevent confusion between old and new classes.}
    \label{fig:mi}
\end{figure}

Despite their promising progress, these existing methods suffer from the \emph{feature confusion} problem, which hampers their performance in knowledge learning and maintenance, especially when a single image contains both old and new class objects. The \textit{feature confusion}  is two-fold, \emph{i.e.}, the inter-session confusion and intra-feature confusion. Firstly, as the categories of different sessions are disjoint and trained separately, it is difficult to establish sharp decision boundaries between old and new classes, resulting in feature confusion between new and old classes, known as the \textit{inter-session confusion}~\cite{masana2022class} problem. This problem is more pronounced in multi-label tasks with multiple targets in a single, leading to a significant performance decline in MLCIL. Secondly, as shown in Fig.~\ref{fig:mi} (a), all existing methods extract an image-level feature for an image during the incremental learning process. However, as a single image may contain multiple classes in MLCIL, the image-level feature will lead to intra-feature confusion during the incremental learning process. For example, in Fig.~\ref{fig:mi} (a), the image contains a \textbf{man} riding his \textbf{bike} behind a \textbf{car}. Extracting an image-level feature for this image will not only mix the features of these three classes but also lead to label conflict~\cite{KRT} as the classes are learned in different incremental sessions, resulting in class confusion and semantic conflict within the image-level feature, \emph{i.e.}, the \emph{intra-feature confusion problem}. 

To bridge this gap, we propose a novel framework for MLCIL, named~\textbf{CL}ass-Independent \textbf{IN}crement~(CLIN). Different from the existing methods that extract an image-level feature for MLCIL, we propose a novel class-independent incremental network (CINet) to extract multiple class-level embeddings for input images containing multiple classes and learn and maintain the knowledge of different classes by constructing class-specific tokens, as shown in Fig.~\ref{fig:mi} (b). The CINet comprises a \textit{class-level cross-attention} module and a \textit{class-independent classifier}. The former module takes class-specific tokens as input, where each token corresponds to a specific class. It leverages attention mechanisms to extract and generate class-level embeddings associated with each class, effectively producing class-level features, and thereby transforming the multi-label problem into multiple single-label problems. Moreover, it can effectively extract local discriminative features and explore correlations between labels as a special Transformer decoder~\cite{q2l}.
Subsequently, the class-independent classifier replaces the previous joint classifier, establishing a one-to-one correspondence between class-level embeddings and class labels. It predicts the presence of the corresponding labels through binary classification, reducing interference in learning between new and old classes. During the incremental learning, we dynamically extend the class-specific tokens and the class-independent classifier to acquire new knowledge and freeze the tokens and classification heads associated with old classes to prevent forgetting. The proposed CINet not only effectively generates and processes class-level features to address the feature confusion problems but also alleviates inter-session confusion by maintaining knowledge for each old class through a small number of parameters independent of the shared model. 

On this basis, to further address the feature confusion problem in MLCIL, we introduce two novel loss functions to optimize the learning of class-specific tokens and class-level embeddings, respectively. Concretely, we address the challenge of applying metric learning to multi-label tasks and propose a contrastive loss on class-level embeddings, \textit{i.e. multi-label contrastive loss}, aimed at minimizing intra-class distances and maximizing inter-class distances.  Our model achieves significant performance improvements on both existing MLCIL protocols~\cite{KRT,ML-SK}. Moreover, our method maintains high performance in a more challenging setting~\cite{rebll2025}, significantly saving storage space.

The key contributions of this paper are summarized as follows:
\begin{itemize}
\item We propose a novel and effective method, CLIN, which utilizes class-level rather than image-level features to address the MLCIL task. Our approach effectively addresses the feature confusion problem and achieves notable performance improvements.
\item We design a novel scalable network, CINet, consisting of a class-level cross-attention module and a class-independent classifier for class-level embedding generation and class-specific knowledge retention.  
\item We propose a multi-label contrastive loss to further mitigate the interaction between old and new knowledge.
\item Our method exhibits competitive performance comparable to the state-of-the-art methods under two established protocols. Furthermore, in a more challenging setting, it surpassed prior SOTA methods by 8.3\% and 4.4\% on different benchmarks, respectively, demonstrating superior robustness.
\end{itemize}

\section{Related Work}
\label{sec:rela}
\subsection{Single-Label Incremental Learning}
 Mainstream single-label incremental learning research can be broadly divided into the following four categories:

\textbf{Rehearsal-based} methods maintain a set of exemplars representing existing data for simultaneous training with new data from the current task. These methods primarily address the challenge of CIL~\cite{ICARL,tao2020topology,podnet} problem. For instance, ER~\cite{ERbase} establishes a memory buffer to store samples from previous tasks for retraining with new data, while DER++~\cite{der+} suggests applying knowledge distillation penalties to the stored data. Moreover, iCaRL~\cite{ICARL} and its variations ~\cite{lucir,podnet,tao2020topology} prevent forgetting by utilizing the herding technique to select exemplars and formulating distinct KD losses. BIC~\cite{BIC} and other methods~\cite{bic2020} incorporate an additional bias correction process to adapt the classification layer.  Recent approaches~\cite{ashok2022class} introduce adaptive aggregation networks or emulate the feature space distribution of Oracle to enhance rehearsal-based methods. CAP~\cite{kd_pr} stabilizes the feature distribution of previously learned classes by generating augmented prototypes and applying multiple distillation constraints, thereby improving unbiased class-incremental recognition performance.

\textbf{Rehearsal-Free} methods represent a
stricter setting, where no samples, prototypes, or class statistics
from previous tasks are stored. Early rehearsal-free methods primarily rely on regularization to preserve previously learned knowledge. Parameter regularization minimizes updates to weights associated with earlier tasks~\cite{EWC,SI,oewc}. EWC~\cite{EWC} constrains important parameters using the Fisher information matrix, while oEWC~\cite{oewc} and SI~\cite{SI} further refine the estimation of parameter importance over time to better maintain stability. In contrast, data regularization retains old knowledge by treating previous models as soft teachers~\cite{LWF}. For example, LWF~\cite{LWF} leverages knowledge distillation to mitigate forgetting without storing samples, enabling continual learning in a rehearsal-free setting. More recently, rehearsal-free methods have shifted from purely regularization-based strategies toward more sophisticated mechanisms. BUILD~\cite{free1_pr} introduces an out-of-distribution (OOD) detection framework that selects pseudo-samples from new-task data to substitute for previously learned classes, enabling fully buffer-free class-incremental learning. DCC-POC~\cite{free2_pr} improves upon single one-class classifier approaches by employing discriminative and comparable parallel one-class classifiers to enhance both separability and consistency across tasks. SSL-CIL~\cite{pr5} further advances this direction by integrating self-supervised pre-training with prototype learning, effectively mitigating catastrophic forgetting through the decoupling of representation drift, feature confusion, and classifier distortion.

\textbf{Architectural-based} methods provide distinct parameters for each task to prevent forgetting. For example, Abati et al.~\cite{PackNet} and Rusu et al.~\cite{PNN} have proposed various strategies to isolate old and new task parameters, including duplicating a new network for each task to transfer previous knowledge through lateral connections. Recent architectural approaches~\cite{der,dytox,slow_pr} combined with the rehearsal methods achieve a better anti-forgetting effect. These methods dynamically expand or prune the network parameters to accommodate the new data at the expense of limited scalability.

\textbf{Prompt-based} methods learn a small set of insertable pre-trained model prompts instead of directly modifying the encoder parameters. For example, L2P~\cite{l2p} creates only a pool of prompts inserted into the model and matches input images to prompts with image-wise prompt queries. DualPrompt~\cite{dualprompt} proposes multiple pools of prompts including G-Prompts and E-Prompts for learning task-invariant and task-specific knowledge. Moreover, S-Prompts~\cite{S-prompts} learns the prompts domain by domain, and incrementally insert the learned prompts into a pool designed for domain-incremental learning. This setting involves learning the same set of classes under covariate distribution task shifts. It is distinct from the CIL setting presented in our paper, where the goal is to learn emerging object classes in new sessions.

\subsection{Multi-Label Stream Learning.} Multi-label stream learning~(MSL)~\cite{2016new,zhu2018new}, also known as dynamic multi-label learning or new label emerging, involves reusing previously learned models and creating new classifiers for newly emerging labels. In instance data streams, aside from the initial training set, subsequent data does not have class labels available. Therefore, detecting and modeling new labels becomes the primary challenge. For example, MLF~\cite{2016new} assumes that new labels are linear combinations of other labels and inherits the relationships between labels through classifiers. MuENL~\cite{zhu2018new} designs detectors based on input features and predicted label attributes and proposes algorithms for updating the classifier. 

MSL can be considered a special form of weakly supervised learning~\cite{zhu2018new}. Since the data stream for new classes is unlabeled, the core focus is on detecting the emergence of new labels and modeling the relationships between new and known labels, rather than on whether known labels are forgotten. This is fundamentally different from the MLCIL setting. MLCIL tasks require learning labeled new classes without forgetting old ones and being able to perform MLC on all previously learned classes. Therefore, MSL approaches cannot be directly applied to MLCIL tasks.

\vspace{-0.2cm}
\subsection{Multi-Label Class Incremental Learning}
Multi-label class-incremental learning~\cite{KRT, csc,rebll2025,zhang2025specifying} aims to continually build a unified classifier capable of integrating newly acquired knowledge while retaining a comprehensive understanding of multiple co-existing objects within an image. Early attempts, such as AGCN and its variants~\cite{AGCN,AGCN2}, leveraged augmented graph convolution to model class co-occurrence relationships and incrementally expand the label graph. PRS~\cite{PRS} and OCDM~\cite{OCDM} improved online sampling strategies to construct memory buffers closer to real data distributions, thereby alleviating imbalance; however, these online-learning-oriented methods perform suboptimally in offline settings~\cite{KRT}.

KRT~\cite{KRT} formally defined the offline multi-label class-incremental learning task and introduced an extensible ICA module with token distillation to preserve and transfer knowledge, effectively mitigating catastrophic forgetting. APPLE~\cite{ML-SK} further proposed a benchmark protocol and framework combining adaptive pseudo-labeling, cluster-based sampling, and a class-attention decoder to alleviate forgetting in MLCIL. Building upon this line of research, we conduct comprehensive comparisons under two offline MLCIL protocols.

More recently, CSC~\cite{csc} addressed task-level partial labeling by reducing false positives arising from overconfident predictions, thereby improving retention of previously learned classes. In parallel, RebLL~\cite{rebll2025} identified the positive–negative imbalance issue in MLCIL and introduced a dual rebalancing framework that jointly tackles data- and loss-level asymmetry, yielding more stable continual multi-label recognition.

\vspace{-0.1cm}
\section{Methodology}

\label{sec:method}

\begin{figure*}[t]
\setlength{\abovecaptionskip}{0 cm} 
\setlength{\belowcaptionskip}{-0.15cm} 
\begin{center}
    \includegraphics[width=0.95\textwidth,height=0.45\textwidth]{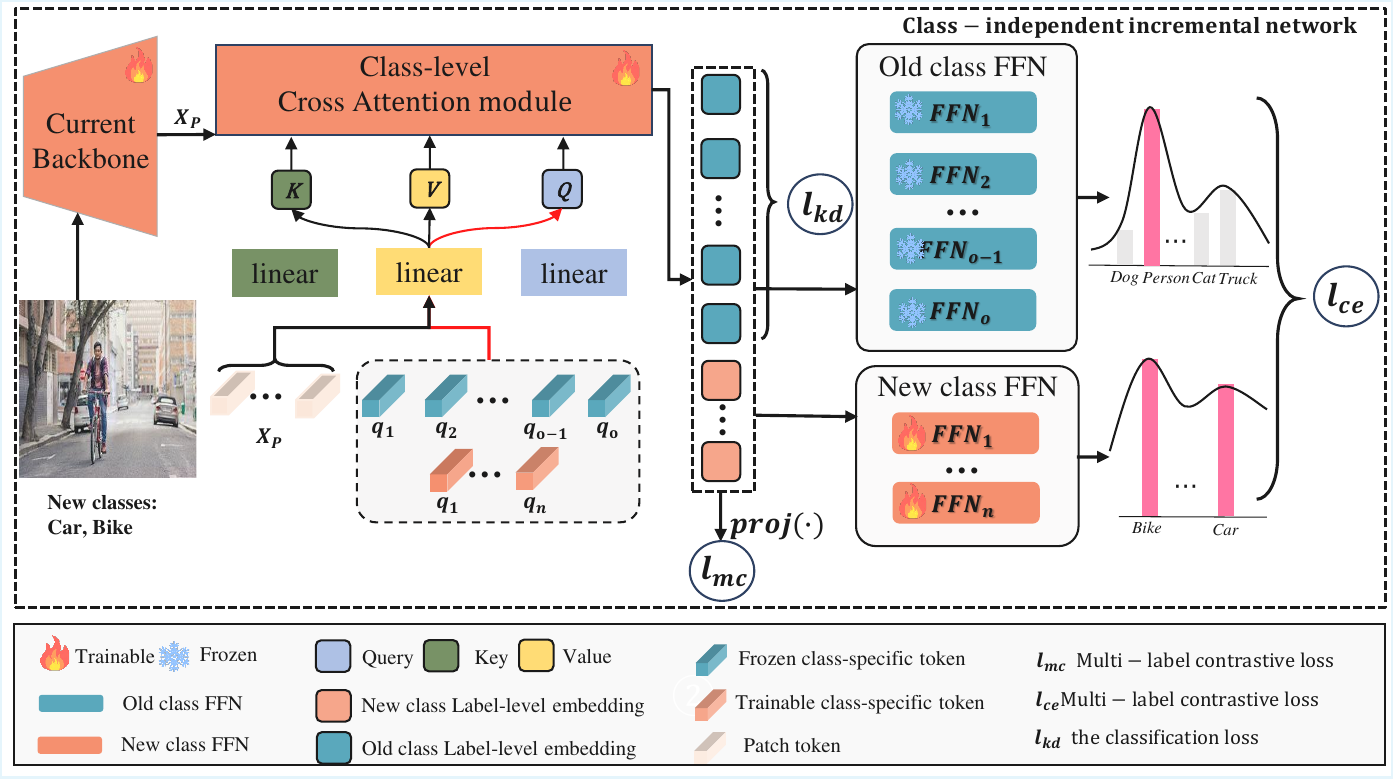}
\end{center}
\caption{The framework of proposed CLIN. During training, the image $x^t$~(In actuality, an input image can contain either one or multiple new classes, and it can also contain one, multiple, or no old classes.) undergoes feature extraction by the backbone and is then converted into patch tokens $x_P$. Together with a set of class-specific tokens, this forms the input for the class-level cross-attention module to generate class-level embeddings $E^t$. The set of class-specific tokens includes trainable tokens $\{\boldsymbol{q}_{c}\}_{c=1}^{C_n}$ for new classes and frozen tokens $\{\boldsymbol{q}_{c}\}_{c=1}^{C_o}$ for old classes. Subsequently, $E^t$ are separately fed into the class-independent classifier to compute $\mathcal{L}_{ce}$ and a projection layer to calculate $\mathcal{L}_{mc}$.}
\label{fig:framework}
\end{figure*}

\subsection{Problem Formulation}
\label{sec:protocol}
The multi-label class-incremental learning problem~\cite{KRT} is defined as follows: suppose we have T sequences of multi-label image sets $ \left\{\mathbf{D}^1,\mathbf{D}^2,...,\mathbf{D}^T \right\}$, where $\mathbf{D}^t=\left\{\mathbf{X}^t, \mathbf{Z}^t\right\}$ is consisted of a training set $\mathbf{X}^t$ and a test set $\mathbf{Z}^t$. Each training set is defined as $\mathbf{X}^t=\left \{ \left ( \mathbf{x}_{i}^{t}, \mathbf{y}_{i}^{t} \right ) \right \}_{i=1}^{N^t}$, where $N^t$ represents the number of training samples, $\mathbf{x}_{i}^{t}$ is the $i$-th training sample and $\mathbf{y}^{t}_{i}\subseteq\mathbf{C}^t$ is a label set with $1\leq |\mathbf{y}^{t}_{i}| \leq |\mathbf{C}^t|$. $\mathbf{C}^t$ denotes the class collection at the $t$-th session and $\forall m,n~(m\neq n),~\mathbf{C}^{m} \cap\mathbf{C}^{n}=\varnothing$\footnote{The training set images from different sessions can be repeated, but the annotated labels are different in~\cite{KRT}, i.e., $ X_m \cap X_n \neq \emptyset$. In protocol~\cite{ML-SK}, image repetition is strictly restricted, i.e., $X_m \cap X_n = \emptyset$.}. In the MLCIL setting, a \textit{multi-label classification} model is required to incrementally learn a \textit{unified} classifier from a sequence of training sessions. At session $t$, only $\mathbf{X}^{t}$ is available during training and the model is evaluated on a combination of test sets $\mathbf{Z}^{1\sim t}=\mathbf{Z}^1\cup\cdots\cup\mathbf{Z}^t$ and is expected to recognize \textit{all} the encountered classes $\mathbf{C}^{1\sim t}=\mathbf{C}^1\cup\cdots\cup\mathbf{C}^t$ present in the test images.

\subsection{Overall Framework}
\label{sec:framework}
Fig.~\ref{fig:framework} illustrates the pipeline of the CLIN approach.
The structure of CLIN comprises two main components: a feature extractor backbone and a class-independent incremental network~(CINet).
The pre-trained feature extractor is represented by $f(\cdot;\theta)$, which is continual training during the incremental learning process. 
The class-independent incremental network comprises a class-level cross-attention module and independent classifiers. The former module is defined as $\gamma(\cdot;\omega)$, with a set of trainable class-specific tokens $\{\boldsymbol{q}_{c}\}_{c=1}^{|\mathbf{C}|}$, where $\boldsymbol{q}_{c} \in \mathbb{R}^{D}$, $D$ denotes the dimension of token and $|\mathbf{C}|$ denotes the number of classes.
The latter module is denoted as $\varphi(\cdot;\phi)$, including $|\mathbf{C}|$ mutually exclusive feed forward network (FFN) classifiers $\varphi(\cdot;\phi_c)$, where $c$ denotes the class ID. 

Assuming the $t$-th incremental session, the input of image $x^t$ from $\mathbf{D}^t$ is fed into the feature extractor $f(\cdot;\theta)$ and projected into the image token $X_P\in \mathbb{R}^{L \times D}$, where $L$ is the token length and $D$ is the feature dimension. Following this, $X_P$ serves as input to the CINet and obtains class-level embeddings $E^t$. In the CINet, the $\gamma(\cdot;\omega)$ extends $|\mathbf{C}^t|$~($C_n$) class-specific tokens (Orange one) to learn new classes and freeze $|\mathbf{C}^{1\sim t-1}|$~($C_o$) old tokens (Blue one) to prevent forgetting of old classes. 

Afterward, class-level embeddings $E^t$ are mapped to a new space through a projection layer $\text{Proj}(\cdot)$, and in this projection space, a multi-label contrastive loss $\mathcal{L}_{mc}$ is introduced to further improve the feature confusion problem. Moreover, similar to previous methods, to alleviate the forgetting of old knowledge, we design a knowledge distillation loss $\mathcal{L}_{kd}$ on class-level embeddings $E^t$ and old class-level embeddings $E^{t-1}$.
Finally, the $E^t$ is fed into the class-independent classifier $\varphi(\cdot;\phi)$ to perform the classification task and calculate $\mathcal{L}_{ce}$ loss. Moreover, we follow the pseudo-labeling methods from previous works~\cite{KRT,ML-SK} to mine old class information from input images. The following sections in this part will provide a detailed description of the CLIN method's structure and loss functions.
\vspace{-0.2cm}
\subsection{The Structure of Proposed CLIN}
Unlike previous methods that leverage image-level representations for the MLCIL task, the CLIN approach proposes a novel class-independent incremental network to extract multiple class-level representations for each image and achieve incremental learning through network expansion and parameter freezing. The structure of CLIN consists of two main components: the feature extractor and class-independent incremental network as shown in Fig.~\ref{fig:framework}.

\vspace{0.1cm}
\noindent\textbf{Feature Extractor.} Given an image $x \in \mathbb{R}^{H \times W \times 3}$ as input, we use a feature extractor to extract image-level features $f \in \mathbb{R}^{h \times w \times c}$, where $h, w$ represent the height and width of the feature map, respectively, and $c$ denotes the dimension of the features. Subsequently, we add a linear projection layer to project the features from dimension $c$ to $D$ to match with the dimension of the class-independent incremental network in the next stage and reshape the projected features to be patch tokens $X_p \in \mathbb{R}^{L \times D}$ where $L=hw$. 

\noindent\textbf{Class-Independent Incremental Network.} To implement and leverage class-level representations, we introduce the class-independent incremental network~(CINet), consisting of two main components: a class-level cross-attention module with a set of trainable class-specific tokens and class-independent classifiers. In this chapter, we first introduce the fundamental structure of the CINet and subsequently elucidate how the CINet incrementally acquires new knowledge while preserving old knowledge in the MLCIL task. 

\vspace{0.1cm}
\textbf{(1) Class-level Cross-Attention:} The first component of the CINet consists of a cross-attention module $\gamma(\cdot;\omega)$ and a set of learnable tokens $\{\boldsymbol{q}_{c}\}_{c=1}^{|\mathbf{C}|}$, where $|\mathbf{C}|$ denotes the number of classes. 

The first input to $\gamma(\cdot;\omega)$ is the patch tokens $X_p$ obtained from the input image via a feature extractor. Moreover, to extract class-level representations and prevent confusion between new and old classes, we design a set of learnable tokens $\{\boldsymbol{q}_{c}\}_{c=1}^{|\mathbf{C}|}$ (denoted as $X_q\in \mathbb{R}^{|\mathbf{C}|\times D}$), with the number of tokens matching the categories. Typically, the cross-attention module consists of two parts: multi-head attention block (MultiAttBlock) and multi-layer perceptron (MLP). It can be formalized as:
\begin{eqnarray}
 & &E_{a} =X_{q}+\mathrm{MultiAttBlock(}(\mathrm{Norm}(X_q,X_p))), \\
 & &E =  E_{a}+\mathrm{MLP}(\mathrm{Norm}(E_a)),
\end{eqnarray}      
where MultiAttBlock($\cdot$) and Norm($\cdot$) denote the cross-attention and layer normalization in \cite{gdtrs}, respectively, and MLP is a multi-layer perceptron with a single hidden layer. 

In the multi-head attention block, we use the class-specific token $X_q$ as the Query ($\mathbf{Q}$). Thus, through the multi-head self-attention mechanism, the model can learn attention weights for specific embeddings of the labels. For example, if an image contains a \textbf{car}, the corresponding embedding is expected to be associated with a high attention weight for the \textbf{car} label. Meanwhile, the patch tokens $X_p$ of the image serve as Key ($\mathbf{K}$) and Value ($\mathbf{V}$). Without loss of generality, it can be formulated as:
\begin{eqnarray}
& & \mathbf{Q} = \mathbf{W}_q X_q=\mathbf{W}_q[q_1,q_2,...q_{|\mathbf{C}|}], \nonumber \\
& & \mathbf{K} = \mathbf{W}_k X_p,  \nonumber\\
& & \mathbf{V} = \mathbf{W}_v X_p, \nonumber\\
& &  \mathbf{z} = \mathbf{W}_o\mathrm{softmax}\left(\frac{\mathbf{Q} \mathbf{K}^{T}}{\sqrt{l/h}}\right)\mathbf{V}+\mathbf{b}_o,
\end{eqnarray}      
where $l$ is the embedding dimension, $h$ is the number of attention heads. 

As the key module of the CINet, the class-level cross-attention module takes an image-level feature as input and outputs its class-level representations, denoted by $E\in \mathbb{R}^{|\mathbf{C}|\times D}$. The class-level embeddings $E$ keep the same dimension as class-specific tokens $X_q$ (i.e. Query), where each row of $E$ corresponds to the embedding of the image under the context of a specific class. Finally, class-level embeddings $E$ are input into the class-independent classifier $\varphi(\cdot;\phi)$ for classification tasks.

\vspace{0.1cm}
\textbf{(2) Class-independent Classifier:}
The input of the class-independent classifier is the class-level embeddings $E=\{e_1,e_2,...e_{|\mathbf{C}|}\}$, where the class-level embedding $e_{c}$ captures the features of the input image in the context of label $c$. To establish the correspondence between class labels and class-level embeddings, and to ensure no interference between different categories, especially between old and new classes, we utilize $|\mathbf{C}|$ separate Feed Forward Networks (FFNs) as classifiers instead of joint classifier. It predicts the presence of corresponding labels through binary classification.

For each label $c$, $\mathrm{FFN}_c$ contains a linear layer and the parameters of $\varphi(\cdot;\phi_c)$ are a weight $w_c \in \mathbb{R}^{D\times1}$ and a bias $b_c$. Specifically, we have a total of $|\mathbf{C}|$ independent FFNs as the classifier of $|\mathbf{C}|$ classes, which can be formulated as:
\begin{equation}
     p_c= \sigma(w_c\times e_{c} + b_c), \,\,  c\in [1,|\mathbf{C}|],    
\end{equation}
where $p_c$ is the prediction probability of $c$-th class, and $\sigma$ is a sigmoid function. 

\vspace{0.1cm}
\textbf{(3) CINet Implements Incremental Learning.}
Assuming the $t$-th incremental session, the input image $x^t$ with its class collection $\mathbf{C}^t$ is initially processed through the feature extractor $f(\cdot;\theta)$ to obtain the patch tokens $X_p \in\mathbb{R}^{L\times D}$, serving as the image input component for the class-level cross-attention module.

As shown in Fig~\ref{fig:framework}, we extend the class-specific token set by introducing vectors $X^t_q\in \mathbb{R}^{|\mathbf{C}^t|\times D}$, where the number of rows in the vectors corresponds to the number of new classes $|\mathbf{C}^t|$, i.e., each row vector represents a class-specific token to learn the knowledge of one new class. Additionally, the class-specific token set also includes already learned tokens for old classes $X^{1\sim t-1}_q\in \mathbb{R}^{|\mathbf{C}^{1\sim t-1}|\times D}$. These tokens are gradually added in the previous sessions. The class-specific tokens $X^t_q$ (Orange) for new classes are concatenated with the old tokens $X^{1\sim t-1}_q$ (Blue) from the previous in the row dimension, forming the input vectors $X_q \in \mathbb{R}^{|\mathbf{C}^{1\sim t}|\times D}$:
\begin{equation}
    \mathbf{X}_{q} = [\textcolor{blue}{\mathbf{X}_{q}^1},...,\textcolor{blue}{\mathbf{X}_{q}^{t-1}},\textcolor{orange}{\mathbf{X}_{q}^t}], \,\, X_q^t=\{\boldsymbol{q}_{c}\}_{|\mathbf{C}^{1\sim t-1}|+1}^{|\mathbf{C}^{1\sim t}|} ,
\end{equation}
where the new tokens $X^t_p$ are trainable while the old tokens $X^{1\sim t-1}_q$ are frozen in the $t$-th training session. The frozen tokens effectively preserve the knowledge of previous classes to mitigate catastrophic forgetting, while simultaneously optimizing the trainable parameters to learn new classes.
Then, the patch tokens $X_p$ and class-specific tokens $X_q$ learn attention weights regarding specific embeddings through a multi-head attention mechanism, obtaining class-level embeddings $E\in \mathbb{R}^{|\mathbf{C}^{1\sim t}|\times D}$, which are then input into the classifier for classification.

To learn new classes, the class-independent classifier also adds $|\mathbf{C}^t|$ Feed Forward Networks (FFNs). Similar to class-specific tokens, FFNs are categorized into two types: frozen FFNs $\varphi(\cdot;\phi_1)$,...,$\varphi(\cdot;\phi_{|C^{1\sim t-1}|})$ and trainable FFNs $\varphi(\cdot;\phi_{|C^{1\sim t-1}|+1})$,...,$\varphi(\cdot;\phi_{|C^{1\sim t}|})$. These serve the purpose of maintaining the old knowledge and learning new knowledge. Ultimately, the image's classification loss $\mathcal{L}_{ce}$ is calculated based on the predicted probabilities from each FFN.
\vspace{-0.3cm}
\subsection{The Loss Function of CLIN}
After introducing the structure of the CLIN method, we proceed to discuss the loss functions that optimize the CLIN method. To address the feature confusion problem, we introduce a novel term in the loss function to optimize the learning of class-level embeddings: multi-label contrastive loss. Additionally, similar to previous approaches, our loss function includes base losses such as classification loss and distillation loss, aiming to achieve the classification task and prevent catastrophic forgetting.

\noindent\textbf{Multi-label Contrastive Loss:}
In contrast to regular supervised learning on independently and identically distributed data, continual learning faces challenges such as poor discriminative ability in non-stationary data distribution. To address this issue, regular supervised learning methods often employ techniques like metric learning~\cite{topic} and contrastive learning~\cite{co2l}. Since image-level features are associated with multiple labels, \textit{directly applying} them to multi-label tasks is challenging. For example, assuming that the image-level repre sentation of an image containing a \textbf{car} must always be close to each other is unreasonable because the \textbf{car} is just one of many objects in these images, possibly occupying only a small part of the image.
However, in our CLIN approach, we tackle this problem by using attention mechanisms to transform image features into class-level features, effectively \textbf{converting} the \textit{multi-label problem} into multiple \textit{single-label problems}. Furthermore, we propose a \textit{multi-label contrastive loss} to enforce the model to learn diverse and discriminative features for new concepts, further addressing the feature confusion problem.

\begin{figure}[h]
    \centering
        \setlength{\abovecaptionskip}{-0.15cm} 
\setlength{\belowcaptionskip}{-0.25cm} 
    \includegraphics[width=\linewidth, height=0.3\textwidth]{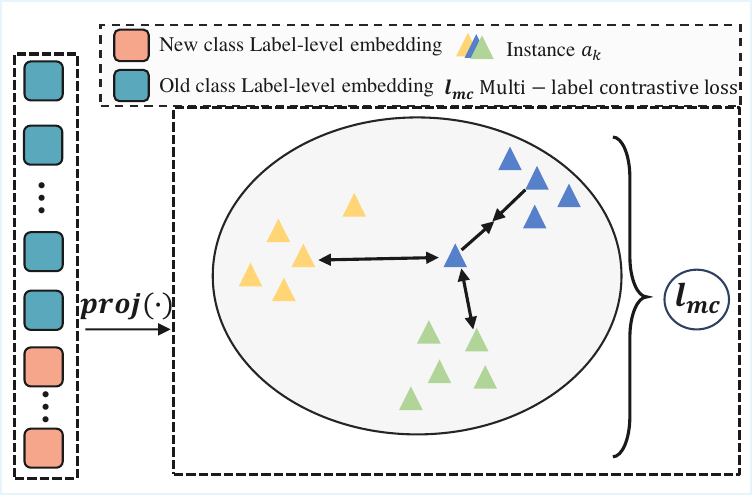} 
    \caption{Detailed illustration of the Multi-label Contrastive Loss ($\mathcal{L}_{mc}$). $\mathcal{L}_{mc}$ is enforced on positive instances to enhance feature discrimination across new and old concepts.}
    \label{fig:mcloss}
\end{figure}

First, after obtaining $E \in \mathbb{R}^{|\mathbf{C}| \times D}$, we use $e_c \in \mathbb{R}^D$ to represent the class-level embedding of the input image in the context of a specific class $c$. Then we introduce a projection layer $\text{Proj}(\cdot)$, which maps $e_c$ to a vector in another embedding space: $z_c = \text{proj}(e_c) \in \mathbb{R}^D$ and we apply the multi-label contrastive loss $\mathcal{L}_{mc}$ in the projected space. 

Specifically, in the $t$-th session, given a small batch \(N\) of images \(X^t = \{x^t_i \in \mathbf{D}^{t} \, | \, i \in \{1, \ldots, N\}\}\) (we omit the superscript \(t\) in the following text as it is unrelated to session). We first input them into the class-level cross-attention module and the projection layer, obtaining class-level embeddings in the mapped space and defining them as the set \(Z = \{z_{ic} \in \mathbb{R}^{D} \, | \, i \in \{1, \ldots, N\}; c \in \{1, \ldots, |\mathbf{C}|\}\}\). Similarly, we define the set of ground truth labels of the minibatch: \(Y = \{y_{ic} \in \{0, 1\} \, | \, i \in \{1, \ldots, N\}; c \in \{1, \ldots, |\mathbf{C}|\}\}\). It's worth noting that we consider the class-level embedding \(z_{ic}\) of an image as an instance rather than the image itself, \(z_{ic}\) is associated with a single true label \(y_{ic}\). We further define \(A = \{\{a_k,y_k\} | \{z_{ic} \in Z \, and \, y_{ic} = 1\}\}\) as the instance set containing class-level embeddings, where \(a_k\) represents the \(k\)-th instance in \(A\), \(y_k\) represents the label of \(a_k\), and \(y_k=c\).

For pairs of class-level embeddings \(a_k\) and \(a_m\) which are randomly selected from the instance set \(A\). Our inter-class objective enforces the following loss on the model:

\begin{equation}\label{eq:angle}
\begin{aligned}
\mathcal{L}_{mc} = \frac{1}{|A|}\sum_{k=1}^{|A|} \sum_{m=1}^{|A|} (1 - cos(a_k, a_m)) * B_{km},
\end{aligned}
\end{equation}
where $\cos(a_k, a_m)$ denotes the cosine similarity between two class-level embedding $a_k$ and $a_m$. \(B_{km}\) represents the relationship of  \(a_k\) and \(a_m\). If $a_k$ and $a_m$ represent the same category, i.e., $y_k = y_m$, then $B_{km} = 1$; otherwise, $B_{km} = -1$. The objective of our \(\mathcal{L}_{mc}\) loss function is to promote feature distinctiveness among different classes, facilitating diversity in class-specific knowledge and clustering features belonging to the same class.

\vspace{0.1cm}
\noindent\textbf{Base Loss:}  The base loss function consists of two portions: (1) the classification loss \(\mathcal{L}_{ce}\), an asymmetric loss~\cite{asl2020}, used to classify the data:
\begin{equation}
\label{eq6}
 \mathcal{L}_{ce}=\frac{1}{N} \sum_{n=1}^{N} \left\{
\begin{aligned}
 & (1-p_c)^{\gamma+}log(p_c), & y_c=1, \\
 & p_c^{\gamma-}log(1-p_c), & y_c=0.
\end{aligned}
\right.
\end{equation}
In the formulation, $\mathbf{y}_c$ represents the binary label indicating whether the image has the label  $c$. The parameters $\gamma+$ and $\gamma-$ are positive and negative focusing parameters, respectively.\\
(2) the distillation loss function \(\mathcal{L}_{kd}\) is applied on the class-level embedding layer to prevent the forgetting of old knowledge:
 \begin{equation}
 \mathcal{L}_{kd} = \|E^{t-1},\Hat{E}^{t}\|,
 \end{equation}
where $E^{t-1}$ is the class-level embeddings of the old backbone, $\hat{E}^{t}$ is the class-level embedding of the old classes in the current backbone ($\hat{E}^{t} = E^{t}[e_1,...,e_{|C^{1\sim t-1}|}]$), and$\|\cdot\|$ represents the $L_2$-norm.

\vspace{0.1cm}
\noindent\textbf{Total Optimization Objective:} For session $t~(t>1)$, our full optimization consists of the two portions: (1) the separation loss includes the multi-label contrastive loss \(\mathcal{L}_{mc}\). (2) the base loss includes the classification loss \(\mathcal{L}_{ce}\) and distillation loss \(\mathcal{L}_{kd}\). We can obtain the total loss function:

\begin{equation}\label{eq:angle}
\begin{aligned}
 \mathcal{L}= \mathcal{L}_{ce}+ \alpha*\mathcal{L}_{mc}\ + \beta*\mathcal{L}_{kd},
\end{aligned}
\end{equation}
where \(\alpha\), \(\beta\) and \(\lambda\) are hyper-parameters. The details are added to the experiments.

\section{Experiment}
\label{sec:experiment}
In this section, we conduct extensive experiments to validate the effectiveness of our algorithm, particularly by evaluating its performance on the MS-COCO~\cite{coco2014} and PASCAL VOC 2007~\cite{voc2007} datasets using two widely recognized benchmark protocols~\cite{KRT,ML-SK}. Additionally, we delve into a series of ablation studies and visualization aimed at assessing the importance of each component, providing deeper insights into the workings of our method. The section is structured as follows: we begin by introducing the experimental setup and implementation details in Sec.~\ref{Experimental detail}. Subsequently, we present the experimental results on the two protocols in Sec.~\ref{main result}. Finally, we introduce the ablation study and provide a comprehensive analysis of our method in Sec.~\ref{Ablation Study}. 

\subsection{Experimental Setup and Implementation Details} 
\label{Experimental detail}
\noindent\textbf{Dataset.} MS-COCO~\cite{coco2014} is a large-scale dataset constructed for segmentation and object detection tasks first and has been widely used for evaluating MLC. It contains 122,218 RGB images of 80 object categories, where each class has 82,081 images in the training set and 40,137 images in the validation set. PASCAL VOC~\cite{voc2007} is a frequently used dataset for MLC and consists of $9,963$ images for 20 object categories, where 5,011 images are in the train-value set, and 4,952 images are in the test set. Our experimental results are reported as the averages over five runs with different random seeds, with variations within ±\textbf{0.3\%}. 

\textbf{Benchmark Protocols.} 
We conduct comprehensive experiments under two standard benchmark protocols and an additional challenging setting.
\noindent \textbf{1). Standard COCO Protocols.}  
Following the protocol in~\cite{KRT}, class names are arranged alphabetically and the training set is partitioned into multiple incremental sessions based on category order. Although different sessions may include the same images, the provided labels differ across increments. For incremental phase division, we adopt settings commonly used in CIL~\cite{lucir,der+}. Specifically, on the MS-COCO dataset, we evaluate our method under the \textbf{B0-C10} and \textbf{B40-C10} protocols, where \(B\) denotes the number of base classes and \(C10\) indicates the addition of 10 new classes per incremental session.
\noindent \textbf{2). VOC Protocols.}  
On the PASCAL VOC benchmark, we follow a similar incremental setup, evaluating our method under the \textbf{B0-C4} and \textbf{B10-C2} protocols. Here, \(B\) and \(C\) denote the number of base and incremental classes, respectively.
\noindent \textbf{3). Challenging COCO Protocols.}  
To further assess robustness under more demanding conditions, we adopt stricter configurations inspired by recent work~\cite{rebll2025}, which introduce either more incremental stages or fewer initial classes. Specifically, on MS-COCO, we evaluate under the \textbf{B20-C4} and \textbf{B0-C5} settings, providing a more rigorous benchmark for incremental learning performance.

\begin{table*}[t]
\centering
\renewcommand\arraystretch{1.2}
\footnotesize

\setlength\tabcolsep{12pt}{
\begin{tabular}{lccccccccc}
\hline
\multirow{3}{*}{\textbf{Method}} & \multirow{3}{*}{\shortstack{\textbf{Buffer}\\\textbf{Size}}} 
& \multicolumn{4}{c}{\textbf{MS-COCO B20-C4}} & \multicolumn{4}{c}{\textbf{MS-COCO B0-C5}} \\
\cline{3-10}
 & & \multicolumn{3}{c}{Last Acc} & Avg.Acc 
   & \multicolumn{3}{c}{Last Acc} & Avg.Acc \\
\cline{3-10} 
 & & mAP & CF1 & OF1 & mAP & mAP & CF1 & OF1 & mAP \\
\hline
Joint & \multirow{2}{*}{-} & 81.8 & 76.4 & 79.4 & - & 81.8 & 76.4 & 79.4 &  \\
FT~\cite{asl2020} &  & 19.4 & 10.9 & 13.4 & 36.5 & 22.5 & 15.0 & 23.6 & 48.1 \\ 
\midrule
ER~\cite{ERbase} & \multirow{4}{*}{5/class}  & 41.9 & 32.9 & 29.8 & 53.0 & 40.1 & 32.9 & 32.3 & 54.6 \\
PODNet~\cite{podnet} &  & 58.4 & 44.0 & 39.1 & 67.7 & 58.2 & 45.1 & 40.8 & 67.2 \\
oEWC~\cite{oewc} & & 58.9 & 45.7 & 39.3 & 68.4 & 59.09 & 46.4 & 44.9 & 69.0 \\
DER++~\cite{der+} &  & 57.3 & 41.4 & 35.5 & 65.5 & 57.9 & 43.6 & 39.2 & 68.2 \\ \midrule
LwF~\cite{LWF}  & \multirow{3}{*}{0} & 34.6 & 17.3 & 31.8 & 55.4 & 50.6 & 36.3 & 41.1 & 66.2 \\
KRT~\cite{KRT} & & 45.2 & 17.6 & 33.0 & 64.0 & 44.5 & 22.6 & 37.5 & 63.1 \\
CSC~\cite{csc} &  & 60.6 & 44.5 & 43.0 & 69.8 & 63.4 & 50.7 & 50.1 & 71.1 \\
RebLL~\cite{rebll2025} &  & \underline{60.1} & 51.3 & 49.2 & 69.2 & 63.5 & 53.5 & 51.9 & 71.7 \\
 \rowcolor{gray!20} \textbf{Ours} &  & \textbf{68.4} & \textbf{67.2} & \textbf{62.9} & \textbf{73.5} & \textbf{67.9} & \textbf{66.2} & \textbf{62.7} & \textbf{73.3} \\
\hline
\end{tabular}}
\caption{Experimental results on MS-COCO dataset under the challenge protocols (B20-C4 and B0-C5).}
\label{tab:challenge}
\end{table*}

\begin{table*}[htb!]
\renewcommand\arraystretch{1.3}

\footnotesize
\setlength\tabcolsep{6pt}
\begin{center}
\setlength{\abovecaptionskip}{-0.1cm} 
\setlength{\belowcaptionskip}{-0.5cm} 
{\begin{tabular}{l|c|c|cccc|cccc}
\hline
\multirow{3}{*}{\textbf{Method}} & \multirow{2}{*}{\textbf{Source}} & \multirow{3}{*}{\textbf{Buffer Size}} & \multicolumn{4}{c|}{\textbf{MS-COCO B0-C10}}  & \multicolumn{4}{c}{\textbf{MS-COCO B40-C10}}    \\ \cline{4-11} 
     &  \multirow{2}{*}{\textbf{Task}}            & & \multicolumn{1}{c|}{Avg. Acc} & \multicolumn{3}{c|}{Last Acc}& \multicolumn{1}{c|}{Avg. Acc} & \multicolumn{3}{c}{Last Acc} \\   \cline{4-11}
     &          & & \multicolumn{1}{c|}{mAP}  &  CF1 & OF1 & mAP    & \multicolumn{1}{c|}{mAP} &  CF1 & OF1 & mAP\\  \midrule 
    Upper-bound & Baseline &    -     &   \multicolumn{1}{c|}{-}  & 76.4  &  79.4      & 81.8   & \multicolumn{1}{c|}{-}   & 76.4  &  79.4   & 81.8   \\   \midrule 
    FT~\cite{asl2020}& Baseline & \multirow{8}{*}{0}   &  \multicolumn{1}{c|}{38.3} & 6.1 & 13.4 & 16.9 ($\downarrow 55.0$)  & \multicolumn{1}{c|}{35.1} & 6.0 & 13.6 & 17.0 ($\downarrow 58.8$)   \\
    PODNet~\cite{podnet} & CIL &     &  \multicolumn{1}{c|}{43.7} & 7.2 & 14.1 & 25.6 ($\downarrow 46.3$)  & \multicolumn{1}{c|}{44.3} & 6.8 & 13.9 & 24.7 ($\downarrow 51.1$)     \\
     oEWC~\cite{oewc} & CIL& &  \multicolumn{1}{c|}{46.9}    &  6.7    & 13.4  & 24.3 ($\downarrow47.6$) & \multicolumn{1}{c|}{44.8} & 11.1 & 16.5  & 27.3 ($\downarrow 48.5$)\\
     LWF~\cite{LWF}& CIL &     &  \multicolumn{1}{c|}{47.9}    &   9.0   & 15.1  & 28.9 ($\downarrow43.0$) & \multicolumn{1}{c|}{48.6} & 9.5 & 15.8 &  29.9 ($\downarrow 45.9$) \\
     AGCN~\cite{AGCN}& MLCIL&     &  \multicolumn{1}{c|}{72.4} &53.9 &56.6 &61.4($\downarrow 10.5$)   &\multicolumn{1}{c|}{73.9}& 58.7& 59.9& 69.1($\downarrow 16.7$)\\
     
     KRT~\cite{KRT} & MLCIL&    &  \multicolumn{1}{c|}{74.6} & 55.6  & 56.5 & 65.9 ($\downarrow 6.0$) & \multicolumn{1}{c|}{77.8}  & 64.4   & 63.4 & 74.0 ($\downarrow 1.8$) \\ 
     
     CSC~\cite{csc} & MLCIL&    &  \multicolumn{1}{c|}{78.0} & \textbf{64.9} & \textbf{66.8} & \textbf{72.8}($\uparrow 0.9$) & \multicolumn{1}{c|}{78.2} &\textbf{ 65.7} & \textbf{67.0} & {75.0}($\downarrow 0.8$)\\

   \rowcolor{gray!20} \textbf{Ours} & MLCIL&    & \multicolumn{1}{c|}{\textbf{78.6}}  &{61.6}  & {61.5} & {71.9 ($\downarrow 0.0$)} &  \multicolumn{1}{c|}{\textbf{79.3}} & {65.6}& {65.4} &{\textbf{75.8} ($\downarrow 0.0$)} \\
  \midrule

    iCaRL~\cite{ICARL} & CIL &   \multirow{9}{*}{20/class}   & \multicolumn{1}{c|}{59.7} & 19.3& 22.8 &  43.8 ($\downarrow 31.2 $)  & \multicolumn{1}{c|}{65.6}  & 22.1& 25.5  & 55.7 ($\downarrow 22.0$)    \\
    
    BiC~\cite{BIC} & CIL &      & \multicolumn{1}{c|}{65.0} &31.0 &38.1 & 51.1 ($\downarrow 23.9$) & \multicolumn{1}{c|}{65.5}   &  38.1 & 40.7 &  55.9 ($\downarrow 21.8$)      \\
    
    ER~\cite{ERbase} & CIL &      & \multicolumn{1}{c|}{60.3} & 40.6 & 43.6 & 47.2 ($\downarrow 27.8$)   & \multicolumn{1}{c|}{68.9}  & 58.6 & 61.1 & 61.6 ($\downarrow 16.1$)     \\
    
    TPCIL~\cite{tao2020topology} & CIL &   &  \multicolumn{1}{c|}{69.4} & 51.7 &52.8 &60.6 ($\downarrow 14.4$) &  \multicolumn{1}{c|}{72.4}  & 60.4   &   62.6     & 66.5 ($\downarrow 11.5$)     \\ 
    
    PODNet~\cite{podnet} & CIL &    &   \multicolumn{1}{c|}{70.0} & 45.2  &48.7 & 58.8 ($\downarrow 16.2$) & \multicolumn{1}{c|}{71.0} & 46.6   & 42.1 &64.2 ($\downarrow 13.5$)    \\
    
    DER++~\cite{der+}   & CIL &     &  \multicolumn{1}{c|}{72.7} & 45.2 &48.7 & 63.1 ($\downarrow 11.9$) &  \multicolumn{1}{c|}{73.6} & 51.5   & 53.5       &  66.3 ($\downarrow 11.4$)       \\
    
    KRT~\cite{KRT} & MLCIL&    & \multicolumn{1}{c|}{76.5}  & 63.9  & 64.7 & 70.2 ($\downarrow 4.8$) &  \multicolumn{1}{c|}{78.3} & 67.9 & 68.9 & 75.2 ($\downarrow 2.5$) \\

     CSC~\cite{csc} & MLCIL&    &  \multicolumn{1}{c|}{79.6} & \textbf{67.8} & \textbf{68.6} & {74.8}($\downarrow 0.2$) & \multicolumn{1}{c|}{78.7} &{68.2} & {69.7} &{76.0}($\downarrow 1.7$)\\
     
     \rowcolor{gray!20}\textbf{Ours} & MLCIL&    & \multicolumn{1}{c|}{\textbf{79.8}}  & {67.0}  & {68.1} & \textbf{75.0 ($\downarrow 0.0$)} &  \multicolumn{1}{c|}{\textbf{80.0}} & \textbf{70.2}& \textbf{71.9} &\textbf{77.7 ($\downarrow 0.0$)} \\
   \hline
\end{tabular}}
\end{center}
\caption{Main class-incremental results on MS-COCO dataset. Compared methods are grouped based on different buffer sizes. Buffer Size 0 means no rehearsal is required, where most SOTA CIL methods are not applicable anymore. All the input size is $224\times224$ and all metrics are in \%. All compared results are obtained from KRT~\cite{KRT}. }
\label{tb:result1}
\end{table*}



\noindent\textbf{Compared Method.} Our method was compared to multiple SOTA continual learning methods following the MLCIL protocol. This included the SOTA MLCIL methods KRT~\cite{KRT}, APPLE~\cite{ML-SK}, RebLL~\cite{rebll2025}, and CSC~\cite{csc}, along with various single-label class-incremental learning methods for comparison. These single-label methods comprised regularization-based approaches such as EWC~\cite{oewc}, LwF~\cite{LWF}, and rehearsal-based methods like ER~\cite{ER}, TPCIL~\cite{tao2020topology}, DER++\cite{der+}, and PODNet\cite{podnet}. Additionally, we considered prompt-based methods L2P~\cite{l2p}, DualPrompt~\cite{dualprompt}, and CODA~\cite{smith2023coda} both utilizing pre-trained ViT-B models and representing the latest SOTA methods in the incremental learning field. The Upper-bound joint trains all classes simultaneously. However, it is not an Incremental Learning (IL) method but serves as an upper bound for performance comparison. 

\textbf{Evaluation metrics.} Following the precedent set by prior incremental learning studies~\cite{KRT,lucir}, we report two key metrics, average accuracy and final accuracy. For a comprehensive evaluation of all learned categories within each session, we utilize the mean average precision (mAP) and report both the average mAP (the mean of the mAP across all sessions) and the final mAP (mAP in the final session). To further enhance the evaluation of performance across all incremental tasks, we include the per-class F1 measure (CF1) and overall F1 measure (OF1) alongside the final accuracy. 

\textbf{Implementation details.} All methods were implemented using PyTorch and trained on 2 RTX 3090 GPUs. We employed a TResNetM backbone pre-trained on ImageNet-21k (L2P~\cite{l2p} and Dual-prompt utilized ViT-B/16 pre-trained on ImageNet-21k as the backbone). The input resolution is set to $h \times w = 224 \times 224$ or $h \times w = 448 \times 448$ depending on the training protocol. The projection dimension $D$ and hidden dimension $l$ of the CINet module were configured as $768$. Our method was trained for 20 epochs using the Adam optimizer and OneCycleLR scheduler, with a weight decay of $1 \times 10^{-4}$. The batch size was set to 64. For training the base model, the learning rate was set to $4 \times 10^{-5}$. During the incremental sessions, we set the learning rate to $1 \times 10^{-4}$ for COCO and $8 \times 10^{-5}$ for VOC. For the hyper-parameters of total optimization, $\alpha$ is typically set to 0.05 and $\beta$ is set to 80. Data augmentation techniques included rand augmentation and cutout. Additionally, we conducted experiments three times and reported the average results. 

\begin{table*}[t!]

\renewcommand\arraystretch{1.3}

\footnotesize
\setlength\tabcolsep{8pt}
\begin{center}
\begin{tabular}{lcccccc}
\hline
\multirow{2}{*}{\textbf{Method}} &\multirow{2}{*}{\textbf{Backbone}} &\multirow{2}{*}{\textbf{Param.}} & \multicolumn{2}{c}{\textbf{MS-COCO B10-C10}}  & \multicolumn{2}{c}{\textbf{MS-COCO B40-C10}}    \\ \cline{4-7} 
     &   &  & \multicolumn{1}{c}{Avg. Acc} & \multicolumn{1}{c}{Last Acc}& \multicolumn{1}{c}{Avg. Acc} & \multicolumn{1}{c}{Last Acc} \\ \midrule
    Upper-bound         & \multirow{9}{*}{ViT-B/16}& \multirow{9}{*}{86.0M}   &  -   &      83.16 &  -& 83.16   \\
    L2P~\cite{l2p}      &  & & 73.28 ($\downarrow6.82$) & 67.72  ( $\nabla$ 15.44 )    & 73.07 ($\downarrow7.50$)       &  70.42   ( $\nabla$ 12.74  )      \\
    L2P-R~\cite{l2p}    &  &  &73.87 ($\downarrow6.23$)  & 68.22  ( $\nabla$ 14.94 )  & 73.64 ($\downarrow6.93$)       &  71.68  ( $\nabla$ 11.48 )          \\ 
    Dual-prompt~\cite{dualprompt} & & & 74.67 ($\downarrow5.42$) & 69.39  ( $\nabla$ 13.77 )   &  74.45 ($\downarrow6.12$)   &   71.95 ( $\nabla$ 11.21  )     \\
    Dual-prompt-R~\cite{dualprompt}    &  & & 74.87 ($\downarrow5.23$) & 70.20 ( $\nabla$ 12.96  )  & 74.54 ($\downarrow6.03$) &  72.60    ( $\nabla$  10.56 )      \\
    CODA~\cite{smith2023coda} & & & 75.84($\downarrow4.26$) &69.87($\nabla$ 5.31) &75.31($\downarrow5.26$) &72.49($\nabla$ 5.31)\\
    
     CODA-R~\cite{smith2023coda} & & & 76.07($\downarrow4.03$) &70.96($\nabla$ 4.22) &75.50($\downarrow5.07$) &72.92($\nabla$ 4.88)\\
           
       \rowcolor{gray!20} \textbf{Ours}   &            & &   \textbf{78.89} ($\downarrow1.21$) &  \textbf{72.24} ( $\nabla$ \textbf{10.92} )   &\textbf{79.62} ($\downarrow0.95$)   &  \textbf{76.24} ( $\nabla$ \textbf{6.92} )   \\ 
       
  \rowcolor{gray!40}  \textbf{Ours-R}&           &  &    \textbf{80.10} ($\downarrow0.00$) & \textbf{75.18} ( $\nabla$ \textbf{7.98} )   &\textbf{80.57} ($\downarrow0.00$)   &    \textbf{77.80} ( $\nabla$ \textbf{5.36} )   \\  

    \midrule
    
\end{tabular}
\end{center}
\vspace{-1.2mm}
\caption{Class-Incremental results (mAP\%) on MS-COCO dataset against prompt-based CIL method. $\nabla$ indicates the gap towards the Upper Bound of the corresponding backbone. All the input size is $224\times224$ and all metrics are in \%.}
\label{tb:l2p}
\end{table*}

\begin{table*}[h]
\centering

\setlength{\belowcaptionskip}{-0.2cm}
\renewcommand\arraystretch{1}
\footnotesize
\setlength\tabcolsep{19 pt}{
\begin{tabular}{lccccc}
\toprule 
\multirow{3}{*}{\textbf{Method}} & \multirow{3}{*}{\textbf{\shortstack{Buffer\\Size}}} & \multicolumn{2}{c}{\textbf{VOC B0-C4}} & \multicolumn{2}{c}{\textbf{VOC B10-C2}} \\ \cline{3-6}
 & & \textbf{Avg. Acc} & \textbf{Last Acc} & \textbf{Avg. Acc} & \textbf{Last Acc} \\ \cline{3-6}
 & & \textbf{mAP (\%)} & \textbf{mAP (\%)} & \textbf{mAP (\%)} & \textbf{mAP (\%)} \\ \midrule
 
\textbf{Upper-bound} & - & - & 93.6 & - & 93.6 \\ \midrule
FT~\cite{asl2020} & \multirow{3}{*}{0} & 82.4 & 62.9 ($\downarrow 24.3$) & 70.1 & 43.0 ($\downarrow 38.2$) \\
KRT~\cite{KRT} & & 89.1 & 80.2 ($\downarrow 7.0$) & 84.3 & 71.5 ($\downarrow 9.7$) \\
\textbf{Ours} & & \textbf{91.1} & \textbf{86.2 ($\downarrow 0.0$)} & {88.7} & {81.2 ($\downarrow 0.0$)} \\ \midrule

iCaRL~\cite{ICARL} & \multirow{8}{*}{2/class} & 87.2 & 72.4 ($\downarrow 14.8$) & 79.0 & 66.7 ($\downarrow 16.8$) \\
BiC~\cite{BIC} & & 86.8 & 72.2 ($\downarrow 15.0$) & 81.7 & 69.7 ($\downarrow 13.8$) \\
ER~\cite{ERbase} & & 86.1 & 71.5 ($\downarrow 15.7$) & 81.5 & 68.6 ($\downarrow 14.9$) \\
TPCIL~\cite{tao2020topology} & & 87.6 & 77.3 ($\downarrow 9.9$) & 80.7 & 70.8 ($\downarrow 12.7$) \\
PODNet~\cite{podnet} & & 88.1 & 76.6 ($\downarrow 10.6$) & 81.2 & 71.4 ($\downarrow 12.1$) \\
DER++~\cite{der+} & & 87.9 & 76.1 ($\downarrow 11.1$) & 82.3 & 70.6 ($\downarrow 12.9$) \\
KRT~\cite{KRT} & & 90.7 & 83.4 ($\downarrow 3.8$) & 87.7 & 80.5 ($\downarrow 3.0$) \\

\rowcolor{gray!20} \textbf{Ours} & & \textbf{92.4} & \textbf{87.2 ($\downarrow 0.0$)} & \textbf{90.3} & \textbf{83.5 ($\downarrow 0.0$)} \\
\bottomrule 
\end{tabular}%
}
\caption{Class-incremental results on PASCAL VOC dataset. Compared methods under different protocols against comparison methods. All the input size is $224 \times 224$.}
\label{tb:result2}
\end{table*}

\subsection{Comparison with State-of-the-arts.}
\label{main result}

\noindent \textbf{Results on Challenge COCO Protocols:} 
To further examine model robustness under more demanding incremental conditions, we evaluate our method on the B20-C4 and B0-C5 challenge protocols, where the number of incremental stages increases and each session introduces fewer new classes. These conditions exacerbate catastrophic forgetting and intensify both intra-feature and inter-session confusion, making them a more suitable stress test for multi-label incremental learning.

As reported in Table~\ref{tab:challenge}, our method achieves the highest performance across all evaluation metrics and significantly surpasses CSC under both challenge configurations. Under B20-C4, the proposed approach achieves a final mAP of 68.4\%, \textbf{exceeding CSC by 7.8\%}. Under B0-C5, our method reaches a final mAP of 67.9\%, \textbf{improving over CSC by 4.5\%}. These substantial margins confirm that the CLIN framework is particularly effective in scenarios where old and new labels frequently co-occur within the same image and where incremental stages provide limited opportunities for model adaptation.

The superior performance arises from three core properties of our design: (1) class-level embeddings that eliminate intra-feature mixing, (2) class-independent classifiers that prevent interference between historical and novel concepts, and (3) a contrastive separation mechanism that explicitly mitigates semantic overlap between old and new label spaces. These results strongly validate that the CLIN framework possesses exceptional anti-forgetting ability and sustained learning capability, proving the effectiveness and robustness of our proposed CINet and dual loss mechanism in challenging MLCIL tasks.






\noindent \textbf{Results on Standard COCO Protocols:}
Table \ref{tb:result1} presents the results of the MS-COCO B0-C10 and B40-C10 benchmark. Our method consistently outperforms almost all comparison methods in terms of average accuracy (mAP) and final accuracy (CF1, OF1, and mAP). Specifically, when the buffer size is larger (20 per class), our method achieves the best final accuracies of \textbf{75.0\%} and \textbf{77.7\%} on the two benchmarks, surpassing SOTA CIL methods by \textbf{11.9\%} and \textbf{11.4\%} respectively. It also outperforms the MLCIL method, KRT, by \textbf{4.8\%} and \textbf{2.5\%}. When the buffer size decreases (5 per class), it shows larger performance gains compared to other continuities. Notably, when the buffer size is set to 0, our method maintains superior performance, far exceeding regularization-based methods, and other rehearsal-based methods, with improvements of \textbf{42.5\%} and \textbf{45.8\%} on the B0-C10 benchmark, respectively. Furthermore, our method is more suitable for rehearsal-free scenarios. Specifically, under the B0-C10 protocol, our method achieves average and final accuracy of \textbf{78.6\%} and \textbf{71.9\%}, respectively, compared to KRT (buffer size=0), representing improvements of \textbf{4.8\%} and \textbf{6.0\%}.

Although CSC achieves slightly higher performance than our method on the Standard COCO benchmarks, exceeding it by approximately \textbf{ 0.9\%} and \textbf{2.2\%}, our approach surpasses CSC by \textbf{7.8\%} and \textbf{4.5\%} under the more demanding Challenge COCO settings. This demonstrates that CLIN exhibits greater stability and superiority in realistic and more challenging incremental learning conditions.


The current SOTA methods in SLCIL are prompt-based approaches. These methods, based on the VIT-B pre-trained model and utilizing prompts, demonstrate powerful capabilities in preventing catastrophic forgetting. For a more comprehensive comparison, we not only evaluate our method on the B40-C10 on the COCO dataset but also the more challenging B0-C10. As prompt-based methods use a different backbone model compared to our approach, we use the tendency towards an upper bound ($\nabla$) to measure the performance of each method given a specific backbone. Table \ref{tb:l2p} illustrates the comparison between our method and prompt-based methods. We observe that our method consistently outperforms prompt-based methods comprehensively, with higher performance across all scenarios, both with and without a buffer, compared to the KRT method. Specifically, our approach shows gaps of \textbf{7.33\%} and \textbf{4.66\%} compared to the upper bound, while DualPrompt exhibits gaps of \textbf{12.96\%} and \textbf{10.56\%} compared to the upper bound. In comparison to the Prompt method, we achieve improvements of at least \textbf{5.63\%}. It is worth noting that despite our method having a lower upper bound, the average accuracy on the two benchmarks is improved by at least \textbf{4.64\%}.

\begin{table*}[t]

\centering 
\footnotesize
\setlength\tabcolsep{16 pt}{
\begin{tabular}{lccccc} 
\hline 
\multirow{3}{*}{\textbf{Methods}} & \multirow{2}{*}{\textbf{Buffer}} & \multicolumn{2}{c}{\textbf{Split-COCO}} & \multicolumn{2}{c}{\textbf{Split-COCO}} \\ 
 & \multirow{2}{*}{\textbf{Size}} & \multicolumn{2}{c}{\textbf{B40-C10}} & \multicolumn{2}{c}{\textbf{B0-C20}} \\ 
\cline{3-6} 
 & & Avg. Acc & Last Acc & Avg. Acc & Last Acc \\ 
\midrule 
Upper bound & - & - & 86.43 & - & 86.43 \\ 
\midrule 
FT & \multirow{2}{*}{0/class} & 35.83 & 11.12 ($\downarrow$ 71.92) & 51.87 & 23.60 ($\downarrow$ 58.93) \\ 
\textbf{Ours} & & \textbf{84.36} & \textbf{82.54} ($\downarrow$ 0.50) & \textbf{83.39} & \textbf{81.71} ($\downarrow$ 0.82) \\ 
\midrule 
iCaRL~\cite{ICARL}& \multirow{8}{*}{20/class} & 76.69 & 65.60 ($\downarrow$ 17.44) & 76.53 & 64.54 ($\downarrow$ 17.41) \\ 
ER~\cite{ERbase} & & 72.30 & 64.05 ($\downarrow$ 18.99) & 63.59 & 50.14 ($\downarrow$ 32.81) \\ 
TPCIL~\cite{tao2020topology} & & 69.41 & 71.20 ($\downarrow$ 11.84) & 73.54 & 68.89 ($\downarrow$ 14.06)\\ 
PODNet~\cite{podnet} & & 77.11 & 66.12 ($\downarrow$ 16.92) & 74.78 & 61.01 ($\downarrow$ 21.94) \\ 
PASS~\cite{pass} & & 73.80 & 59.44 ($\downarrow$ 23.60) & 72.16 & 49.88 ($\downarrow$ 33.07) \\ 
DER++~\cite{der+} & & 66.71 & 55.77 ($\downarrow$ 27.27) & 73.82 & 67.33 ($\downarrow$ 15.62) \\ 
APPLE~\cite{ML-SK} & & 82.05 & 74.61 ($\downarrow$ 8.43) & 83.49 & 76.65 ($\downarrow$ 6.30) \\ 
\rowcolor{gray!20} \textbf{Ours} & & \textbf{84.79} & \textbf{83.04} ($\downarrow$ 0.00) & \textbf{83.92} & \textbf{82.95} ($\downarrow$ 0.00) \\ 
\hline 
\end{tabular}} 
\caption{Experimental results (mAP\%) of our method and comparison methods on Split-COCO datasets. All the input size is $448\times448$ and all metrics are in \%. All compared results are obtained from APPLE~\cite{ML-SK}} 
\label{tab:result3_1} 

\end{table*}
\begin{table*}[t] 

\centering 

\footnotesize
\setlength\tabcolsep{16 pt}{
\begin{tabular}{lccccc} 
\hline 
\multirow{3}{*}{\textbf{Methods}} & \multirow{2}{*}{\textbf{Buffer}} & \multicolumn{2}{c}{\textbf{Split-VOC}} & \multicolumn{2}{c}{\textbf{Split-VOC}} \\ 
 & \multirow{2}{*}{\textbf{Size}} & \multicolumn{2}{c}{\textbf{B10-C5}} & \multicolumn{2}{c}{\textbf{B0-C5}} \\ 
\cline{3-6} 
 & & Avg. Acc & Last Acc & Avg. Acc & Last Acc \\ 
\midrule 
Upper bound & - & - & 94.18 & - & 94.18 \\ 
\midrule 
FT & \multirow{2}{*}{0/class} & 74.82 & 60.09 ($\downarrow$ 32.26) & 74.10 & 59.73 ($\downarrow$ 30.53) \\ 
\textbf{Ours} & & \textbf{94.66} & \textbf{91.98} ($\downarrow$ 0.37) & \textbf{93.88} & \textbf{89.45} ($\downarrow$ 0.81) \\ 
\midrule 
iCaRL~\cite{ICARL}& \multirow{8}{*}{5/class} & 90.78 & 87.07 ($\downarrow$ 5.28) & 88.33 & 84.78 ($\downarrow$ 5.48)\\ 
ER~\cite{ERbase} & & 86.01 & 73.91 ($\downarrow$ 18.44) & 82.68 & 68.31 ($\downarrow$ 21.95)\\ 
TPCIL~\cite{tao2020topology} & & 90.19 & 84.18 ($\downarrow$ 8.17) & 87.95 & 79.38 ($\downarrow$ 10.88) \\ 
PODNet~\cite{podnet} & & 90.35 & 86.35 ($\downarrow$ 6.00) & 87.78 & 84.12 ($\downarrow$ 6.14)\\ 
PASS~\cite{pass} & & 86.01 & 76.93 ($\downarrow$ 15.42) & 75.58 & 51.84 ($\downarrow$ 38.42)\\ 
DER++~\cite{der+} & & 90.22 & 83.95 ($\downarrow$ 8.40) & 88.95 & 84.82 ($\downarrow$ 5.44)\\ 
APPLE~\cite{ML-SK} & & 91.68 & 89.36 ($\downarrow$ 2.99) & 89.52 & 85.62 ($\downarrow$ 4.64)\\ 
\rowcolor{gray!20} \textbf{Ours} & & \textbf{94.85} & \textbf{92.35} ($\downarrow$ 0.00) & \textbf{95.62} & \textbf{90.26} ($\downarrow$ 0.00)\\ 
\hline 
\end{tabular}} 
\caption{Experimental results (mAP\%) of our method and comparison methods on Split-VOC datasets. All the input size is $448\times448$ and all metrics are in \%. All compared results are obtained from APPLE~\cite{ML-SK}} 
\vspace{-0.2cm} 
\label{tab:result3_2} 
\vspace{-0.2cm} 
\end{table*}

\begin{table*}[t]
\renewcommand\arraystretch{1.3}
\footnotesize
\begin{center}
{

\setlength\tabcolsep{7pt}
{

{
\begin{tabular}{cccccccccccccc}
\toprule
\multirow{2}{*}{Baseline} & \multirow{2}{*}{CINet} & \multirow{2}{*}{MC loss} & \multicolumn{8}{c}{Acc (mAP\%) in each session} &\multicolumn{1}{c}{Avg.} \\
  \cline{5-12} & & & 1 & 2 & 3 & 4 & 5 & 6 & 7 & 8  & \multicolumn{1}{c}{Acc} & \\
\midrule
\checkmark & & & 92.5& 79.5  & 73.0 & 67.8 & 62.7  & 62.6  & 59.9 & 58.4  ($\uparrow$ 0.0)& 69.5  \\
\checkmark & \checkmark & & 92.6 & 83.0 & 79.4  & 77.7  & 76.4  & 76.1  & 74.6 & 73.7 ($\uparrow$ 15.3) & 79.2   \\
\checkmark & \checkmark  & \checkmark & \textbf{93.0} & \textbf{83.4} & \textbf{79.9} & \textbf{78.3} & \textbf{77.0} & \textbf{76.7}  & \textbf{75.5} & \textbf{75.0} ($\uparrow$ 16.6) & \textbf{79.8} \\
\bottomrule
\end{tabular}
}
\caption{The effectiveness of CLIN's different components in performance improvements.}
\label{tb:abalation}
}}
\end{center}
\end{table*}

\noindent\textbf{Result on VOC benchmark:} 
Table \ref{tb:result2} summarizes the experimental results on the VOC benchmark. We observe similar trends as seen in the COCO dataset results. Specifically, our method achieves a significant improvement in the final mAP compared to KRT, with an increase of \textbf{3.8\% (83.4\% \(\xrightarrow{}\) 87.2\%)} and \textbf{3.0\% (80.5\% \(\xrightarrow{}\) 83.5\%)}, surpassing other rehearsal methods by at least \textbf{11.1\%}. It is noteworthy that KRT experiences a significant drop in accuracy on the more challenging benchmark (B10-C2) without a buffer, while our method maintains good accuracy.

\vspace{0.1cm}
\noindent\textbf{Results on Split-COCO and Split-VOC benchmark:} 
Table~\ref{tab:result3_1} and Table~\ref{tab:result3_2} summarize the experimental results on the Split-COCO and Split-VOC benchmarks. We observe that our method exhibits significant improvements compared to rehearsal-based methods on these new protocols and achieves higher advancements compared to the SOTA method APPLE in the field of MLCIL. Specifically, our method outperforms APPLE by \textbf{8.43\%} and \textbf{6.30\%} on the two benchmarks in Split-COCO and achieves a maximum improvement of \textbf{4.64\%} on the relatively simpler Split-VOC. Furthermore, even with a buffer size of 0, our method experiences a maximum accuracy drop of only \textbf{0.78\%}, providing further evidence of the effectiveness of our approach. 

We observe that the CLIN method achieves significant performance gains on the MS-COCO benchmark, yet the relative margin of improvement is less pronounced on PASCAL VOC. We attribute this to the differential suitability between the core mechanism of our method and the specific characteristics of the two datasets. COCO is characterized by high density and complex scenes, which lead to severe Intra-Feature Confusion—the very problem where CLIN's class-level decoupling strategy is designed to be maximally effective. In contrast, VOC scenes are simpler and feature lower object density, resulting in less intrinsic feature entanglement. In these low-confusion scenarios, the advantage of our fine-grained, Token-based decoupling is diminished, suggesting that CLIN is optimally suited for tackling high-feature-confusion problems in complex multi-label environments.

\vspace{-0.1cm}
\subsection{Ablation Study}
\label{Ablation Study}
\noindent\textbf{Ablation of Different Components:} 
To demonstrate how each component of the proposed CLIN framework contributes to performance improvements, we conducted ablation studies by building four models on MS-COCO under the B0-C10 benchmark:
(1) We introduced a distillation loss as the \textbf{baseline} model. (2) We added a class-independent incremental Network~(CINet) to improve the baseline model. (3) We added both the CINet and the proposed multi-label contrastive loss $\mathcal{L}_{mc}$ to improve the baseline model. (4) We added all three components to improve the baseline model.

Based on the results presented in Table~\ref{tb:abalation}, we conduct an ablation analysis on the key components of the proposed CLIN method. Specifically, the \textbf{Baseline model} yielded the lowest final mAP value of \textbf{58.4\%} in Session 8, confirming the challenge of catastrophic forgetting. The incorporation of the \textbf{CINet} module significantly mitigated this issue, boosting the final mAP by \textbf{15.3\%} to \textbf{73.7\%} and achieving an overall average mAP of \textbf{79.2\%}. Furthermore, integrating the \textbf{Multi-label Contrastive Loss ($\mathcal{L}_{mc}$)} further optimized performance. The final, comprehensive model, which includes the Baseline, CINet, and $\mathcal{L}_{mc}$, achieved the highest average mAP of \textbf{79.8\%} and the best final session mAP of \textbf{75.0\%}. Notably, the introduction of $\mathcal{L}_{mc}$ alone improved the average mAP by \textbf{0.6}\textbf{$\%$ ($79.8\%-79.2\%$)} and the final mAP by \textbf{1.3}$\%$ ($75.0\%-73.7\%$). These compelling experimental results collectively demonstrate that the two core components, CINet and $\mathcal{L}_{mc}$, are highly effective in stabilizing performance and preventing forgetting in Multi-label Class-Incremental Learning (MLCIL) tasks.
\begin{figure}
\setlength{\abovecaptionskip}{-0.1cm}
\setlength{\belowcaptionskip}{-0.3cm}
    \centering
    \includegraphics[width=\linewidth]{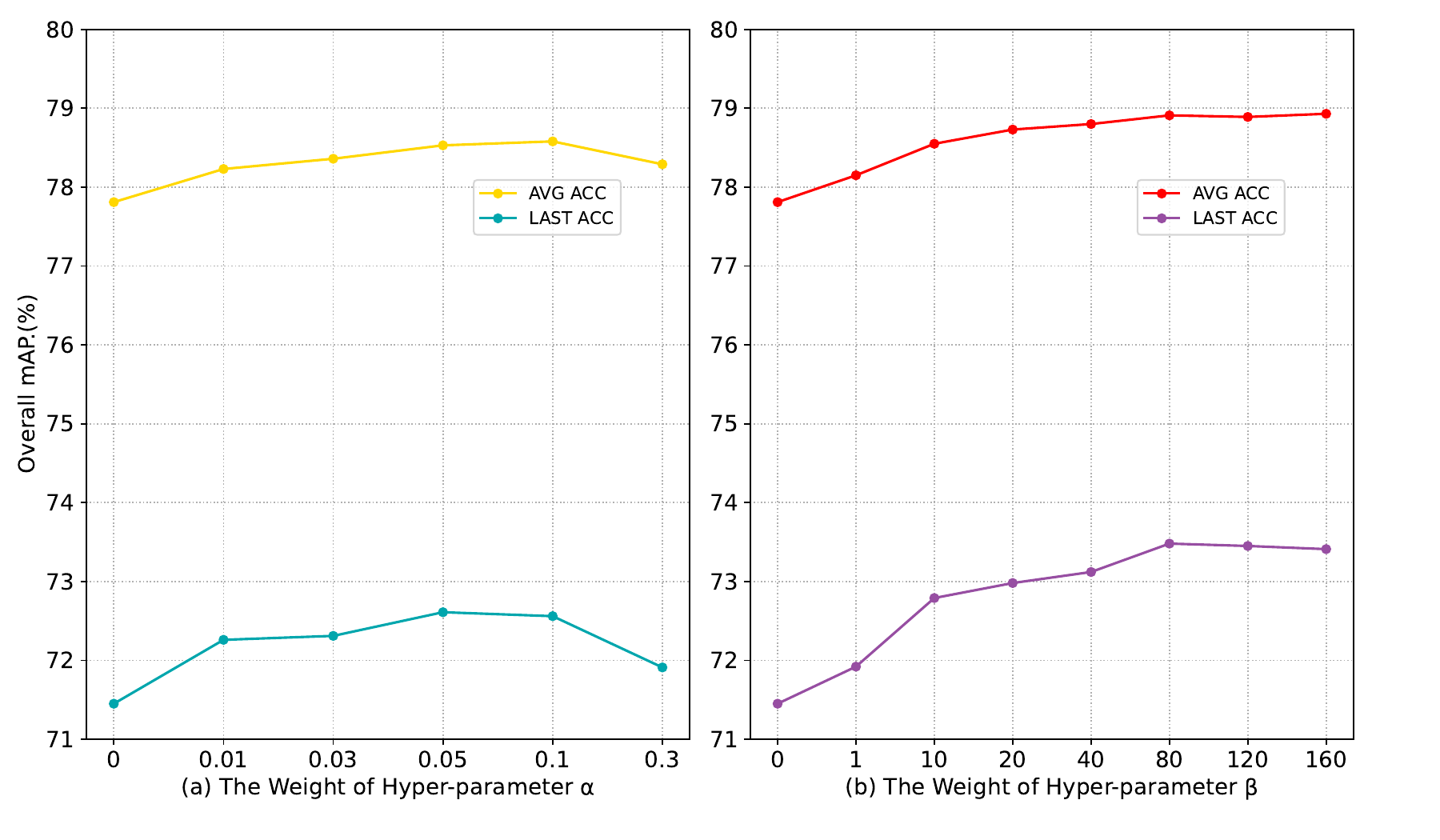}
    \caption{Comparison of the average and last accuracy with different settings of parameters $\alpha$ and $\beta$.}
    \label{fig:sensitivity}
\end{figure}

\noindent\textbf{Sensitive Study of Hyper-parameter:} The hyper-parameters $\alpha$ and $\beta$ control the strength of $\mathcal{L}_{mc}$ and $\mathcal{L}_{kd}$  to together control the total optimization objective. 

We first conducted a sensitivity study on $\alpha$ using the COCO B0-C10 benchmark, where we changed the value of $\alpha$ in the reasonable range of \{0, 0.01, 0.03, 0.05, 0.1, 0.3\}, respectively. The results are shown in Fig~\ref{fig:sensitivity}, and we observed that with the increase of $\alpha$, the final accuracy and average accuracy consistently improved, reaching the optimal final accuracy of \textbf{72.56\%} at $\alpha$=0.05, representing a relative improvement of \textbf{1.11\%}. Subsequently, as $\alpha$ continued to increase, the rate of performance improvement gradually decreased. The experimental trends align with expectations. In contrast to single-label classification, label correlation is crucial in multi-label classification. For instance, when we observe a \textbf{tennis ball} in an image, we can assume a high probability of the \textbf{tennis racket} appearing in the image as well. Therefore, categories in the embedded space should not be overly separated.

To validate the robustness and effectiveness of the CLIN approach, we conducted experiments with different weights of hyper-parameter $\beta$ on C0C0 B0-C10. Specifically, we tested $\beta$ values of 0, 1, 10, 20, 40, 80, 120, and 160, respectively. The comparative results are shown in Fig.~\ref{fig:sensitivity}. We observed that our method achieved the best final performance of \textbf{73.47\%} when $\beta$ was set to 80. Furthermore, our method demonstrated a small performance range across different $\beta$ parameters, ranging from \textbf{71.49\%} to \textbf{73.42\%}, confirming the model's robustness to variations in the $\beta$ parameter. At the same time, we observed that even without distillation loss, our CLIN framework achieved a minimum final accuracy of \textbf{71.49\%}, with only a relative decrease of \textbf{1.93\%}. This demonstrates the effectiveness of our structural design in mitigating catastrophic forgetting.

\begin{table*}[t]
\renewcommand\arraystretch{1.6} 
\setlength{\belowcaptionskip}{-0.2cm} 
\setlength\tabcolsep{20pt} 
\footnotesize 
\centering

\begin{tabular}{lcccccc}
\hline
\multirow{2}{*}{Method} &  & \multicolumn{5}{c}{$M_{per}$} \\ \cline{3-7}
                        &   & 0    & 5     & 10    & 20    & 30    \\ \hline
\multirow{2}{*}{PODNet*~\cite{podnet}} 
                        & Last Acc & 25.6  & 53.4  & 55.6  & 58.8  & 59.0  \\
                        & Avg Acc  & 43.7  & 65.7  & 66.8  & 70.0  & 70.3  \\ \hline
\multirow{2}{*}{KRT~\cite{KRT}} 
                        & Last Acc & 65.9  & 68.3  & 69.4  & 70.2  & 71.5  \\
                        & Avg Acc  & 74.6  & 75.8  & 76.4  & 76.5  & 77.3  \\ \hline
\multirow{2}{*}{Ours}  
                        & Last Acc & \textbf{71.7} & \textbf{72.9} & \textbf{73.8}  & \textbf{75.0}  & \textbf{75.3} \\
                        & Avg Acc  & \textbf{78.4} & \textbf{78.9} & \textbf{79.3}  & \textbf{79.8}  & \textbf{80.0} \\ \hline
\end{tabular}
\caption{Comparison of different methods on MS-COCO with various numbers of exemplars $M_{per}$ per class.}
\label{mper}
\end{table*}

\begin{figure}[t]
\setlength{\abovecaptionskip}{-0.2cm} 
\setlength{\belowcaptionskip}{-0.3cm}
    \centering %
    \includegraphics[width=\linewidth]{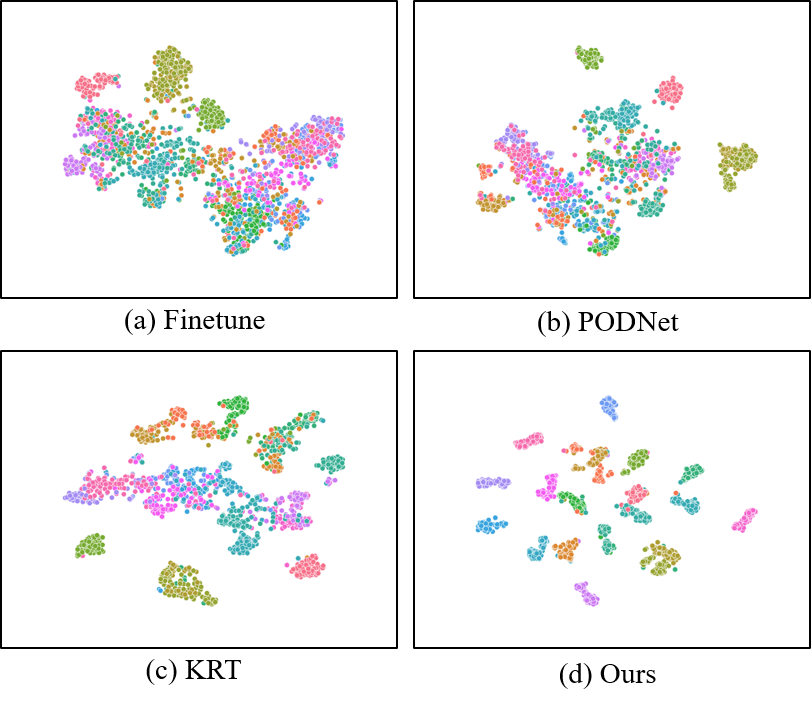} 
    \caption{Comparison of t-SNE visualizations between other methods and our approach, where each color represents a category.}
    \label{fig:sensitivity}
\end{figure}

\vspace{0.1cm}
\noindent\textbf{The Influence of Buffer Size:}
Incremental Learning methods typically utilize additional memory space to store a fixed number of exemplars per old class, denoted as~($M_{per}$), to mitigate catastrophic forgetting. Intuitively, retaining more exemplars may enhance recognition performance but also come with a greater memory burden. Therefore, it is valuable to compare our approach against the rehearsal-based methods PodNet~\cite{podnet} and the SOTA method KRT~\cite{KRT} with different numbers of exemplars in the MS-COCO with B0-C10 setting (refer to Table~\ref{mper}). 

We observe that our method consistently outperforms PodNet* and KRT, achieving a final accuracy of \textbf{75.0\%}. As the number of exemplars (\(M_{per}\)) decreases, the advantage of our method continues to grow. Specifically, with an ample buffer of \(M_{per}=30\), our method outperforms KRT by \textbf{3.8\%} in final accuracy. Moreover, in the scenario with no old examples (\(M_{per}=0\)), our method exhibits a remarkable improvement of \textbf{5.8\%}. This implies that CLIN does not rely on the storage of old examples. It is noteworthy that when the $M_{per}=0$, the accuracy of the rehearsal-based method is seriously attenuated and our method significantly outperforms PodNet* by \textbf{46.1\%(25.6\%$\xrightarrow{}$71.7\%)} on last accuracy.


\vspace{0.1cm}
\noindent\textbf{Visualization of CLIN Method:} To demonstrate that the CLIN approach effectively addresses category confusion issues, we conducted t-SNE visualizations on the final models of various methods under the COCO B0C10 protocol, focusing on the earliest learned 0-20 classes. Through visualization, the observations are as follows: (a) Finetune methods exhibit severe category confusion, leading to significant catastrophic forgetting issues. (b,c) The visualizations of the Rehearsal-based method PODNet and NE-based method KRT show some boundaries between categories compared to the Finetune method. However, the features of different classes remain highly confusing. This indicates that global or task-level features alone cannot resolve confusion issues, resulting in catastrophic forgetting problems. (d) Our CLIN method, through the CINet and proposed loss functions, effectively resolves category confusion in multi-label class incremental learning. The visualization demonstrates clear distinctions between different classes, validating the effectiveness and rationality of our approach.

\section{Conclusion}
\label{sec:conclusion}
This paper primarily investigates the problem of multi-label class-incremental learning. We propose an effective Continual Class-Independent (CLIN) framework to address the challenges of inter-session confusion and the unique intra-feature confusion associated with MLCIL tasks. Firstly, the proposed CINet combines attention mechanisms with scalable class-specific tokens. It effectively generates and processes class-level features to address confusion problems while preserving knowledge for each old class through a small number of parameters independent of the shared model. Secondly, the CINet transforms the multi-label problem into a single-label problem. On this basis, two loss functions are proposed to optimize class-specific tokens and class-level embeddings. In terms of experiments, the performance of the CLIN method consistently outperformed other SOTA methods, particularly in scenarios where no buffer storage was used. We also conducted detailed ablation experiments and visual analyses to demonstrate the effectiveness of the proposed modules and loss functions.
\vspace{0.1cm}

\textbf{Limitation and Feature work.} 
This paper explores and addresses the task of MLCIL from the perspective of incremental learning and has achieved promising results. In terms of label correlation, it attempts to use a Transformer decoder to extract local discriminative features for different labels and establish co-occurrence relationships between categories.

However, it does not delve into the contribution of label dependencies in preventing forgetting and learning new classes. Moreover, due to the insensitivity to algorithm time complexity in incremental learning tasks, our paper does not discuss it.

In the future, we will further (1) investigate the co-occurrence relationship between labels in the query~((token))-to-label structure, using label dependencies as knowledge retention to prevent forgetting and to enhance multi-label classification performance. (2) Investigate the impact of unlearned potential categories in images on the task. (3) Explore the generalizability of our feature decoupling strategy across diverse datasets, particularly analyzing the performance nuances on fine-grained and small-scale benchmarks like PASCAL VOC, and subsequently extend the approach to more network frameworks, such as large-scale natural language models like CLIP, and other multi-label application scenarios.
\clearpage
{
    \small
    \bibliographystyle{ieeenat_fullname}
    \bibliography{main}
}


\end{document}